\documentclass[11pt]{article}

\usepackage[final]{acl}

\usepackage{times}
\usepackage{latexsym}

\usepackage[T1]{fontenc}

\usepackage[utf8]{inputenc}

\usepackage{microtype}

\usepackage{inconsolata}

\usepackage{graphicx}

\usepackage{booktabs}
\usepackage{multirow}
\usepackage{amsmath}
\usepackage[table,dvipsnames]{xcolor}
\usepackage[skins,breakable]{tcolorbox}
\usepackage{makecell}
\usepackage{amssymb}

\title{SwarmBench: Can Large Language Models Act as Agent Swarm Orchestrators?}

\author{
  \textbf{Jinshan Gao\textsuperscript{1}},
  \textbf{Zhuoran Jin\textsuperscript{1,2,}}\thanks{Corresponding authors.},
  \textbf{Tianyi Men\textsuperscript{1,2}},
  \textbf{Kang Liu\textsuperscript{1,2}},
  \textbf{Jun Zhao\textsuperscript{1,2,}}\footnotemark[1]
\\
\\
  \textsuperscript{1}The Key Laboratory of Cognition and Decision Intelligence for Complex Systems, \\ Institute of Automation, Chinese Academy of Sciences, Beijing, China,\\
  \textsuperscript{2}School of Artificial Intelligence, University of Chinese Academy of Sciences, Beijing, China,
\\
  \small{
    gaojinshan2026@ia.ac.cn
  } 
  \small{
      \{zhuoran.jin, tianyi.men, kliu, jzhao\}@nlpr.ia.ac.cn
  }
}

\begin{document}
\maketitle
\begin{abstract}
Large language model-based multi-agent systems are evolving from fixed interaction topologies toward dynamically orchestrated Agent Swarms. However, existing benchmarks are still largely based on single-agent or general-purpose agent tasks, making it difficult to systematically evaluate key orchestration capabilities. We propose SwarmBench, a benchmark that evaluates model performance from multiple perspectives, including accuracy, efficiency, cost, and process quality. Experimental results show that current models exhibit substantial differences in orchestration capability. These differences are reflected not only in final accuracy, efficiency, and cost, but also in the overall quality of the orchestration process itself. Based on these findings, we further propose SwarmExp, a simple yet effective method based on experience extraction and experience replay, which consistently improves the orchestration performance of large language models. Code is available at \href{https://github.com/ying1973/SwarmBench}{https://github.com/ying1973/SwarmBench}.

\end{abstract}

\section{Introduction}
\label{sec: Introduction}

In recent years, large language model-based multi-agent systems (LLM-MAS) have evolved from fixed collaboration paradigms toward more dynamic orchestration paradigms. Early LLM-MAS systems typically relied on predefined roles and communication structures, such as debate, SOP-driven pipeline, or manager-worker collaboration \cite{wu2024autogen, li2025flow, hong2024metagpt}. These systems are relatively easy to control and analyze, but predefined collaboration limits their scalability across tasks. 
In contrast, recent \textbf{\textit{Agent Swarm}} \cite{team2026kimi} further shifts the focus from “designing a fixed team” to “allowing models to organize a team dynamically”. In such systems, a main agent is responsible for decomposing subtasks, creating subagents, coordinating their parallel execution, and finally integrating their outputs. As a result, the key challenge shifts to whether large language models can act as orchestrators that efficiently organize a dynamic swarm to solve complex tasks \cite{dang2026multi}.

However, existing benchmarks are still insufficient for evaluating this emerging paradigm. \textbf{On the one hand, studies on MAS continue to rely on benchmarks originally designed for single-agent systems or general-purpose agents} \cite{mialon2024gaia, jimenez2024swe, merrill2026terminal}, which cannot systematically expose key capabilities of Agent Swarm.
On the other hand, when some tasks appear suitable for multi-agent collaboration \cite{wong2025widesearch}, \textbf{existing evaluations mainly focus on final answer quality while paying less attention to the orchestration process itself}, including the rationality of task decomposition, the effectiveness of subagent creation, the completeness of result aggregation, and so on. Consequently, we still lack a benchmark specifically designed for Agent Swarm orchestration.

To address these limitations, we propose \textbf{SwarmBench}, a benchmark specifically designed to evaluate whether large language models can function as Agent Swarm orchestrators. Specifically, we divide tasks into three categories according to the orchestration capabilities they primarily evaluate: task decomposition, subagent creation and delegation, and subagent result aggregation. 
We construct a dataset consisting of 8 tasks and 400 samples, covering tasks such as root cause analysis, treasure hunt game, wide search, and so on \cite{team2026kimi}.
We further evaluate models from multiple aspects: \textit{accuracy}, \textit{efficiency}, and \textit{cost}. In addition, we introduce a process quality evaluation framework to conduct fine-grained analysis of orchestration quality across three dimensions: task decomposition, subagent delegation, and result aggregation.

Using SwarmBench, we conduct extensive experiments on a variety of proprietary and open-source models. 
The results show that existing models have highly uneven capability distributions. Stronger models not only achieve higher accuracy, but are also more capable of converting parallel structures into real efficiency gains and achieving better Pareto frontiers in the cost-performance trade-off.
Further analysis reveals that different tasks consistently expose different orchestration bottlenecks, among which aggregation is often the most difficult capability to stabilize. Moreover, the study of cost and efficiency shows that system performance indispensably depends on the effectiveness of the orchestration structure, rather than having a simple linear relationship with cost or parallelism.
Based on these findings, we further propose SwarmExp, a simple yet effective method that extracts skill, trick, and model card from trajectories and reinjects them into the main agent's context. Experimental results show that SwarmExp consistently improves performance across multiple representative tasks.

We summarize the contribution as follows.
\begin{itemize}
  \setlength{\topsep}{0pt}   
  \setlength{\itemsep}{2pt} 
  \setlength{\parskip}{0pt}
\item We propose SwarmBench, a benchmark designed to evaluate large language models as Agent Swarm orchestrators, providing a more comprehensive characterization of model capability in swarm scenarios.
\item We conduct extensive experiments and analysis on SwarmBench, revealing the capability structure of current models as orchestrators, showing that the key factor is whether the orchestration structure itself is effective.
\item Based on these findings, we propose SwarmExp, a simple yet effective experience-driven method that improves orchestration capability by extracting experience from trajectories.
\end{itemize}

\section{SwarmBench}
\label{sec: Swarm Bench}
\subsection{Definition of Agent Swarm}
\label{subsec: Definition of Agent Swarm}
Agent Swarm is a hierarchical multi-agent system in which an orchestrator coordinates a dynamically created set of subagents to accomplish a given task. In this paper, we view Agent Swarm as a specific form of multi-agent workflow characterized by three properties: (1) hierarchical orchestration, where a single orchestrator is responsible for task planning, subagent creation, coordination, and final synthesis; (2) dynamic and heterogeneous subagent instantiation, where subagents are created at runtime with different roles, backbone models, tools, and local contexts according to task requirements; and (3) parallel subagent execution, where multiple subagents can execute different subtasks concurrently and report their results back to the orchestrator.

Formally, given an input task $x \in \mathcal{X}$, an Agent Swarm process can be represented as
$$
\mathcal{S}(x) = \bigl(o, \mathcal{H}, \Pi_o, \Pi_h, \mathcal{U}, B \bigr)
$$
where $o$ denotes the orchestrator; $\mathcal{H}$ denotes the set of available subagent configurations; $\Pi_o$ denotes the orchestration policy; $\Pi_h$ denotes the subagent execution policy; $\mathcal{U}$ denotes the available tools and external environments; and $B$ denotes the execution budget, including step limits, agent count, latency, or monetary cost.

A key distinction between Agent Swarm and static or predefined multi-agent workflows is that the concrete subagent set is not fixed before execution. Instead, at global step $t$, the orchestrator observes the current global state $s_t$ and outputs an orchestration action:
$$
a_t^o \sim \Pi_o(\cdot \mid s_t)
$$
where $a_t^o$ belongs to the action space $\mathcal{A}_o.$ When $a_t^o = \texttt{spawn agent}$, the orchestrator creates a new subagent $h_i$ and specifies its configuration
$$
\phi_i = (r_i, m_i, \tau_i, c_i)
$$
where $r_i$ is the role description of the agent; $m_i$ is the backbone model assigned to the agent; $\tau_i \subseteq \mathcal{U}$ is the set of tools available to the agent; and $c_i$ is the local context or subtask prompt of the agent.

In addition, Agent Swarm encourages parallel execution among agents. Let $H_t$ denote the set of currently running agents. The system parallelism can then be defined as $\rho_t = |H_t|$. When an execution process contains stages with $\rho_t > 1$, and the task benefits from parallelizable subtasks, we consider the system to exhibit swarm behavior.

\subsection{Dataset Construction}
\label{subsec: Dataset Construction}

\begin{figure*}[t]
  \includegraphics[width=\linewidth]{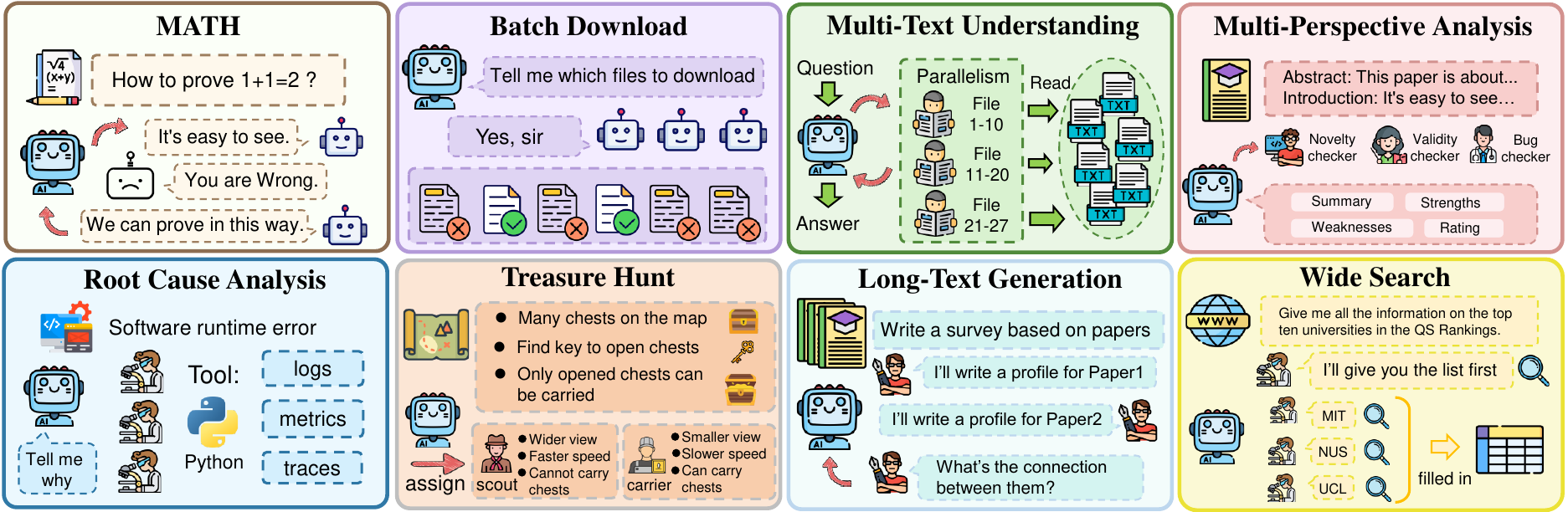}
  \caption{Overview of SwarmBench, a benchmark designed to evaluate LLMs as Agent Swarm orchestrators.}
  \label{fig: swarmbench}
\end{figure*}

SwarmBench aims to evaluate the core capabilities required for large language models to function as Agent Swarm orchestrators, including task decomposition, dynamic subagent creation and delegation, and subtask result aggregation. To achieve this, we design 3 task categories with 8 task types to comprehensively evaluate the key capabilities of the model. For details, see Figure \ref{fig: swarmbench} and Appendix \ref{sec:appendix-Data}.

\textbf{Decomposition-Oriented Tasks.} This category evaluates the ability of the main agent to decompose complex objectives into multiple subtasks that can be executed in parallel or progressively. We include three tasks in this category: \textbf{Batch Download (BD), Multi-Text Understanding (MTU), and MATH}. Batch Download is fully constructed from scratch. Each sample contains a virtual environment and a file download requirement, with an average of \textbf{11.44} files per sample. Multi-text Understanding is built on the Loong \cite{wang2024leave} dataset. We select tasks suitable for multi-agent collaboration, resulting in samples with over \textbf{200K} average context length and more than \textbf{24} documents per sample. MATH is constructed from Omni-MATH \cite{gao2025omni}, using samples with difficulty above \textbf{7}. The system is required to decompose complex problems into verifiable intermediate steps to derive the final solution.

\textbf{Delegation-Oriented Tasks.} This category evaluates the ability of the main agent to select appropriate backbone models, create suitable subagent roles, and delegate different subtasks to corresponding agents according to task requirements. We include three tasks in this category: \textbf{Multi-Perspective Analysis (MPA), Root Cause Analysis (RCA), and Treasure Hunt Game (TH)}. Multi-Perspective Analysis is based on DeepReview \cite{zhu2025deepreview}. The system is required to create subagents with different perspectives such as novelty, effectiveness, and completeness, and organize their collaboration to generate the final review comments. Root Cause Analysis is based on the OpenRCA dataset \cite{xu2025openrca} and requires the system to infer failure causes from different evidence sources including logs, metrics, and traces. Treasure Hunt is a newly constructed task in which the system must dynamically create and delegate subagents with two different roles to explore an unknown map, search for keys, and open chests in an interactive environment.

\textbf{Aggregation-Oriented Tasks.} In this category, subagents typically produce partial drafts, local results, or intermediate content, while the main agent must further filter, integrate, and unify these outputs into a final result with consistent style and complete structure. We include \textbf{Long-Text Generation (LTG) and Wide Search (WS)} in this category. Long-text Generation is based on LongInOutBench \cite{zhang2025lost} and requires the system to integrate content from multiple papers generated by different subagents and produce coherent long-form survey articles. Wide Search is built based on the WideSearch dataset \cite{wong2025widesearch}. Subagents are required to retrieve key information through real-world web search, while the main agent must aggregate the collected information and complete the final answer table.

\subsection{Agent Swarm Framework}
To evaluate the capability of different models as orchestrators, SwarmBench designs a lightweight Agent Swarm framework and applies it across all tasks. In this framework, the main agent is responsible for task decomposition, subagent creation, task assignment, and result aggregation, while subagents independently execute delegated subtasks within their local contexts.

To reduce the influence of the main agent’s own knowledge on task outcomes, the main agent is not allowed to call tools or answer tasks independently, and part of the task context is hidden from the main agent. The backbone model of the main agent is provided by the evaluated model, while subagent backbone models are selected from a configurable model pool. Each model in the pool is associated with a model description written from publicly available information, and the main agent dynamically selects suitable models during execution.
Under this setting, the main agent is needed to create suitable subagent roles based on subtasks, and assign appropriate backbone models to different subagents. Detailed information about the backbone model pool is provided in Appendix \ref{sec:appendix-Model}.

\begin{table*}
  \centering
  \small
  \begin{tabular}{lccccccccc}
    \toprule
    \textbf{Model} & \textbf{MATH} & \textbf{BD} & \textbf{MTU} & \textbf{MPA} & \textbf{RCA} & \textbf{TH} & \textbf{LTG} & \textbf{WS} & \textbf{AVG.}  \\
    \midrule
    \multicolumn{10}{c}{\cellcolor{gray!20}\textbf{Proprietary Models}} \\
    GPT-5.4 & \textbf{82.97} & \textbf{87.84} & \textbf{48.82} & 77.17 & \textbf{10.00} & \textbf{10.33} & \underline{48.19} & \textbf{38.67} & \textbf{50.49} \\
    Claude-Sonnet-4-6 & 60.00 & \underline{61.50} & \underline{46.90} & \textbf{81.43} & \underline{5.50} & 2.38 & \textbf{58.74} & \underline{31.67} & \underline{43.51} \\
    Gemini-3-flash & \underline{73.46} & 55.92 & 31.29 & 78.04 & 3.80 & 6.29 & 30.73 & 29.56 & 38.63 \\
    Doubao-seed-1-8 & 60.00 & 50.77 & 36.12 & 65.39 & 0.00 & \underline{8.00} & 32.19 & 29.87 & 35.29 \\
    Claude-Haiku-4-5 & 44.00 & 34.30 & 40.36 & \underline{78.24} & 2.00 & 6.00 & 40.50 & 19.05 & 33.05 \\
    \midrule
    \multicolumn{10}{c}{\cellcolor{gray!20}\textbf{Open-source Models}} \\
    Kimi-k2.5 & 57.14 & 36.48 & 22.53 & \underline{76.52} & \textbf{8.82} & \textbf{8.33} & 40.67 & 12.05 & 32.81 \\
    Deepseek-v3.2 & 56.00 & \textbf{79.45} & \underline{32.08} & \underline{76.52} & 5.00 & \underline{1.08} & \underline{42.58} & \underline{30.36} & \underline{40.38} \\
    Qwen3.5-397b-a17b & \underline{64.00} & 67.13 & 29.50 & 75.51 & \underline{7.05} & 0.79 & \textbf{45.83} & \textbf{35.20} & \textbf{40.62}   \\
    GLM-5.1 & 61.00 & \underline{71.36} & \textbf{39.00} & \textbf{78.17} & 4.00 & 0.00 & 39.56 & 29.32 & 40.30 \\
    Qwen3-30b-a3b & \textbf{69.00} & 21.70 & 21.56 & 50.71 & 0.00 & 0.00 & 26.93 & 11.32  & 25.15\\
    \midrule
    \multicolumn{10}{c}{\cellcolor{gray!20}\textbf{Single Models}} \\
    GPT-5-mini & 20.00 & \textbf{45.66} & \underline{18.20} & \textbf{68.56} & 1.02 & \textbf{2.13} & \textbf{33.43} & 6.32 & \textbf{24.41} \\
    Claude-Haiku-4-5 & \textbf{31.56} & 40.11 & 16.10 & \underline{65.42} & \textbf{3.19} & \underline{0.50} & 26.19 & 7.01 & \underline{23.76}  \\
    Gemini-2.5-flash-lite & 10.33 & 36.55 & 13.60 & 56.65 & 0.00 & 0.00 & 18.89 & 6.97 & 17.87 \\
    Qwen3.5-35b-a3b & 21.55 & 35.99 & 15.33 & 58.10 & 0.00 & 0.00 & 25.13 & 5.10 & 20.15 \\
    GLM-4.5-air & \underline{24.13} & 33.54 & 17.00 & 60.23 & \underline{1.03} & 0.00 & \underline{26.71} & \underline{7.91} & 21.31  \\
    Doubao-seed-1-6-flash & 14.00 & \underline{41.33} & \textbf{20.10} & 64.71 & 0.00 & 0.00 & 14.00 & \textbf{8.13} & 20.28 \\
    \bottomrule
    
  \end{tabular}
  \caption{\label{table: main results}
    Accuracy evaluation results. Bold indicates the best result, and underline indicates the second-best result. "Proprietary Models" and "Open-source Models" are results where models act as Agent Swarm orchestrators, while "Single Models" refers to single-agent execution results. 
  }
\end{table*}

\section{Experiments}
\label{sec: Experiments}

\subsection{Setup}
\label{subsec: Setup}
\textbf{Models.} We evaluate a comprehensive set of models from multiple sources. Proprietary models are from OpenAI \cite{GPT_5.4}, Anthropic \cite{Claude_Sonnet_4.6}, Google \cite{Gemini_3_flash}, and ByteDance \cite{Doubao_Seed1.8}, while open-source models cover Qwen \cite{Qwen}, Deepseek \cite{DeepSeek}, Kimi \cite{team2026kimi} and GLM \cite{GLM-5}. In all experiments, the main agent models run under the same Agent Swarm framework and share the same subagent backbone model pool. The tool interfaces and output formats of each task are predefined and kept fixed across all models. We also include the backbone models from the subagent pool as single-agent baselines.

\textbf{Metrics.} We report results from three aspects: accuracy, efficiency, and cost. For accuracy, we use the standard evaluation metric of each task; details can be found in Appendix \ref{sec:appendix-Metrics}. Efficiency is measured by the time gain from parallelism. Cost is measured by the actual monetary cost incurred during a complete execution process.

\subsection{Main Results}

Accuracy results can be found in Table \ref{table: main results}. Overall, GPT-5.4 is the strongest and most balanced orchestrator, achieving the best performance on six out of eight tasks, demonstrating stable capability in decomposition, delegation, and aggregation. Claude-Sonnet-4-6 is the closest proprietary competitor. Its strengths are more concentrated on tasks such as MPA and LTG, which rely more heavily on role organization and content synthesis.
In contrast, the capabilities of open-source models are more fragmented and exhibit clear task-specific strengths. Kimi-k2.5 is the most stable model on delegation-oriented tasks, especially RCA and TH. Qwen3.5-397b-a17b performs better on aggregation-oriented tasks such as LTG and WS, showing stronger capability in result integration. Other open-source models, including Deepseek-v3.2 and GLM-5.1, achieve relatively stable performance on tasks such as BD and MTU, but do not demonstrate a consistent orchestration advantage across tasks. Meanwhile, limited by model scale, Qwen3-30b-a3b performs noticeably weaker on delegation- and aggregation-related tasks.
The agreement analysis between LLM-based scores and human evaluation for open-ended task metrics is provided in Appendix~\ref{subsec:appendix-Agreement Between LLM Judge and Human Evaluation}.


\subsection{Cost-Accuracy Trade-off Analysis}
\label{subsec: Cost-Accuracy Trade-off Analysis}
\begin{figure*}[t]
  \includegraphics[width=\linewidth]{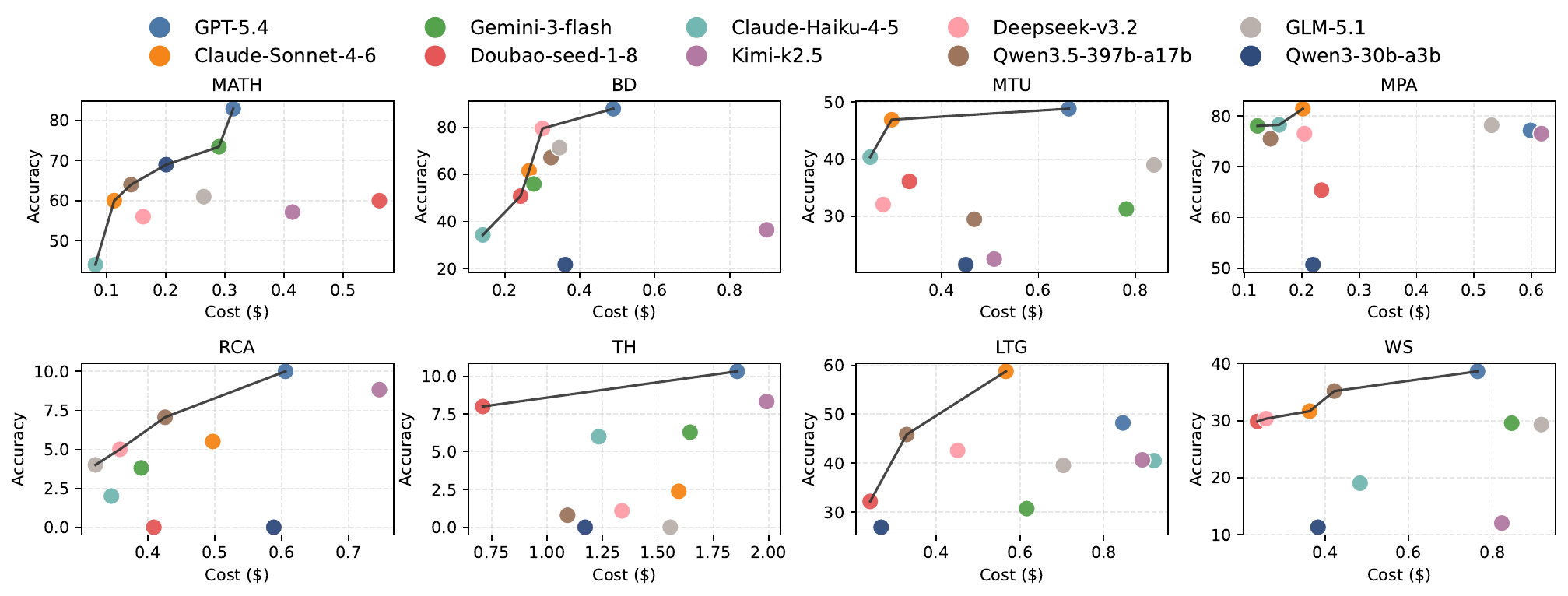}
  \caption{Accuracy-cost Pareto frontier plots across eight tasks.}
  \label{fig: acc_cost_map}
\end{figure*}

To further analyze the cost behavior of models, we measure system accuracy and average cost on all tasks and plot the accuracy-cost curves. We additionally connect the Pareto frontier within each task to illustrate the optimal performance boundary under different budgets, as shown in Figure \ref{fig: acc_cost_map}.

Results show that cost is not a linear function of performance. From the task perspective, performance on MATH and BD generally improves as cost increases. In contrast, on tasks such as MPA, LTG, MTU, RCA, and WS, higher cost does not necessarily lead to better performance, and some lower-cost settings already achieve near-optimal results. From the model perspective, larger models often achieve better cost-performance trade-offs in complex collaborative tasks. Comparing two models from the same family, Qwen3.5-397b-a17b more frequently lies on the Pareto frontier across multiple tasks, while Qwen3-30b-a3b often incurs a higher cost but achieves worse performance.

These observations suggest that cost and performance are not simply linearly related, but largely depend on orchestration quality. Strong orchestrators can achieve better performance with lower cost through more efficient decomposition, delegation, and aggregation, while weaker orchestrators often require more interactions and redundant execution without necessarily obtaining better results.

\subsection{Parallelism and Time Efficiency}
\label{subsec: Parallelism and Time Efficiency}

\begin{figure*}[t]
  \includegraphics[width=\linewidth]{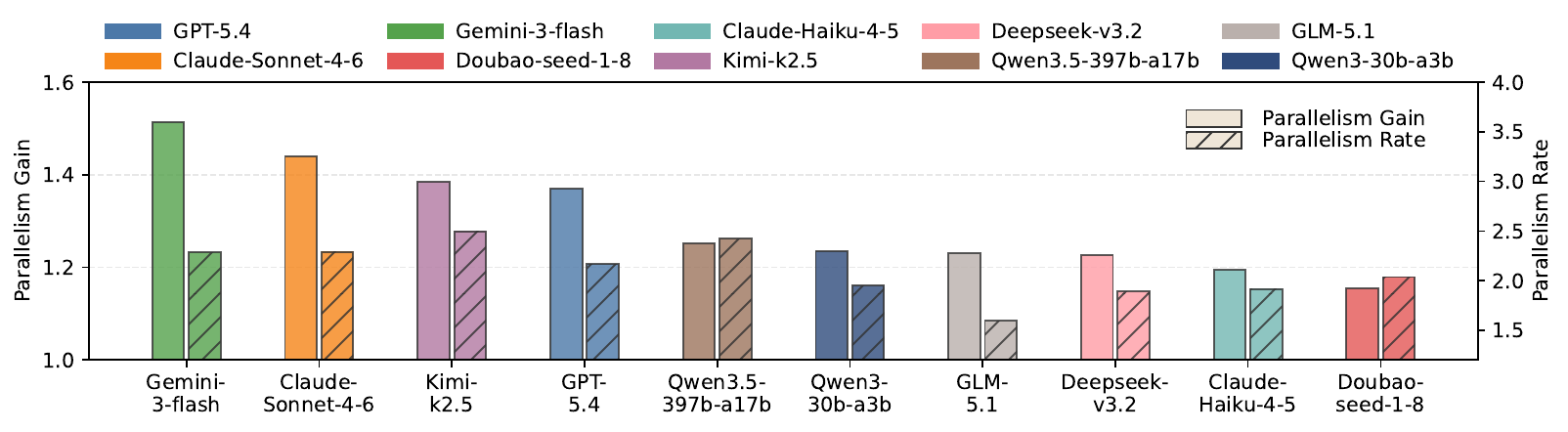}
  \caption{Parallelism gain and rate across ten models. Parallelism gain quantifies the latency reduction brought by parallel execution. Parallelism rate represents the average count of subagents running concurrently in task execution.}
  \label{fig: parallelism_time_map}
\end{figure*}

To analyze execution efficiency, we introduce two metrics related to parallel execution. \textbf{\textit{Parallelism Rate}} measures the average number of concurrently running subagents during execution. \textbf{\textit{Parallelism Gain}} is defined as the ratio between serialized execution time and actual execution time, measuring the latency reduction brought by parallel execution. The former reflects the degree of parallel execution, while the latter reflects its actual efficiency gain. Detailed formulas are provided in Appendix \ref{sec:appendix-Metrics}.

The results in Figure \ref{fig: parallelism_time_map} show that parallelism gain is not simply determined by the amount of parallelism itself. 
In terms of parallelism rate, Kimi-k2.5 and Qwen3.5-397b-a17b significantly outperform other models, indicating that they can more easily learn and activate parallel execution through context prompting. However, a higher parallelism rate does not necessarily lead to higher parallelism gain. While Kimi-k2.5 achieves the highest parallelism rate, its parallelism gain does not surpass Claude-Sonnet-4-6, whose parallelism rate is slightly lower. In contrast, Gemini-3-flash achieves the highest parallelism gain with a moderate parallelism rate. 
These results suggest that the determinant of time reduction is not how many subagents run simultaneously, but whether the main agent constructs an effective parallel structure. Specifically, effective orchestration requires reducing redundant waiting, avoiding repeated execution, and organizing parallel subtasks into genuinely beneficial collaborative workflows.

\subsection{Analysis of Orchestration Quality}

\begin{figure*}[t]
  \includegraphics[width=0.50\linewidth]{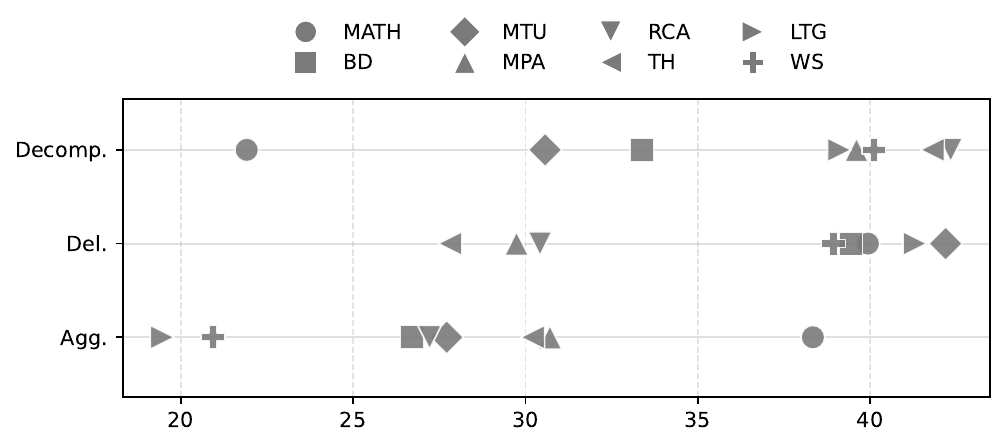} \hfill
  \includegraphics[width=0.50\linewidth]{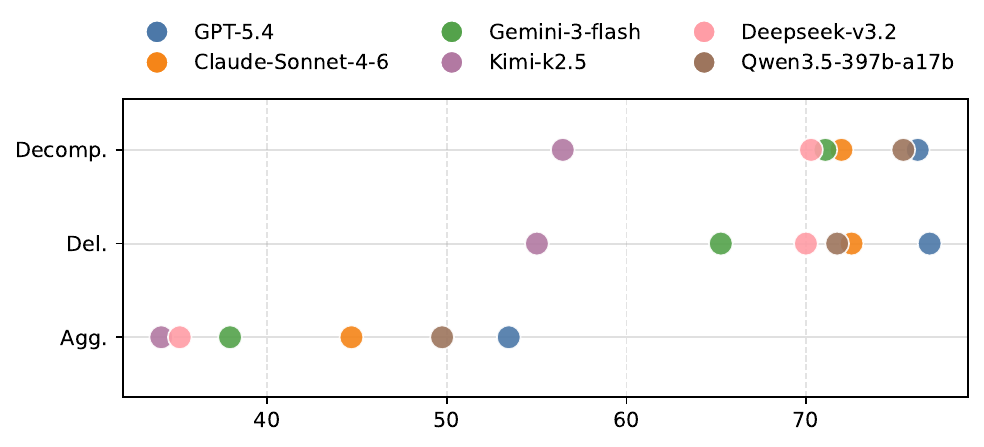}
  \caption {LLM-Judged Orchestration Quality Across Tasks and Models. This figure provides both task-centric and model-centric views of decomposition, delegation, and aggregation performance.}
  \label{fig: llm_judge_score}
\end{figure*}

To further characterize process quality, we introduce an LLM-based evaluation framework that scores the main agent along three orchestration dimensions: task decomposition (Decomp.), subagent creation and delegation (Del.), and subagent result aggregation (Agg.). These dimensions respectively capture whether the main agent can break the task into meaningful subtasks, assign suitable subagents with clear responsibilities, and integrate local outputs into a coherent final response. As shown in Figure \ref{fig: llm_judge_score}, we average the scores across tasks and models to obtain both task-centric and model-centric views of orchestration quality. The agreement analysis between LLM-based process scores and human evaluation is provided in Appendix~\ref{subsec:appendix-Agreement Between LLM Judge and Human Evaluation}.

The task-centric results show that the LLM-based process scores are broadly aligned with our task taxonomy. BD, MTU, and MATH exhibit lower scores in decomposition, with MATH showing the greatest difficulty, indicating that these tasks primarily stress the ability to transform complex objectives into tractable intermediate steps. MPA, RCA, and TH are more challenging in delegation, reflecting their reliance on role construction, model selection, and subtask assignment. In contrast, LTG and WS show the largest drop in aggregation, suggesting that synthesizing multiple local outputs into a unified final answer is the central bottleneck for aggregation-oriented tasks.

The model-centric results reveal that orchestration quality differs not only in overall strength but also in balance across dimensions. GPT-5.4 achieves the strongest and most stable performance, while Claude-Sonnet-4-6 and Qwen3.5-397b-a17b form a competitive second tier with different profiles: the former is more balanced, whereas the latter is stronger in decomposition and delegation. Weaker models tend to suffer especially in aggregation, and the overall score distribution also shows that aggregation is consistently lower than the other two dimensions. This suggests that current models are comparatively better at initiating and assigning subtasks than at reliably integrating subagent outputs into a coherent final decision.

\section{SwarmExp: Experience-Driven Swarm Orchestration}
\label{sec: SwarmExp}
\subsection{Definition}

\begin{figure}[t]
  \includegraphics[width=\linewidth]{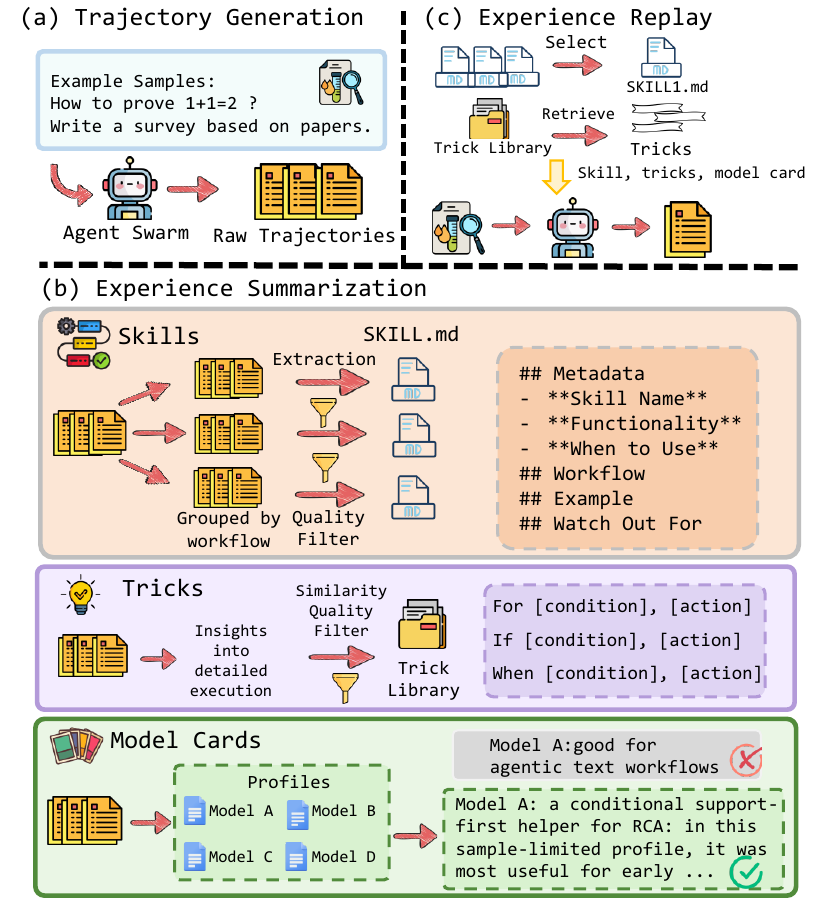}
  \caption{Overview of SwarmExp, consisting of three stages: trajectory generation with the base Agent Swarm, experience summarization into skills, tricks, and model cards, and experience replay by injecting experience back into the main agent.}
  \label{fig: SwarmExp}
\end{figure}

Experiments on SwarmBench show that the limitations of existing models lie not only in final answer quality, but also in the orchestration process, including unstable task decomposition, unreasonable subagent delegation, and weak result integration. Based on these observations, we argue that reusable orchestration experience can be accumulated through task execution \cite{zhang2026evoroute, zhang2026g, hao2026pushinglimitsllmtool}. By extracting and reusing such experience, orchestration capability can be improved. Based on this idea, we propose SwarmExp, a simple yet effective experience-driven enhancement method for Agent Swarm. Figure \ref{fig: SwarmExp} illustrates the framework.

SwarmExp follows a simple two-stage pipeline. First, we run inference on SwarmBench using the initial Agent Swarm framework and collect the full trajectories of the model. We then summarize these trajectories and extract three types of reusable knowledge: skill, trick, and model card.

\textbf{Skill.} This component focuses on high-level workflow patterns \cite{jiang2026xskill, alzubi2026evoskill, si2026context, li2026aee}. We use LLMs to summarize orchestration patterns from trajectories and organize them into candidate SKILL.md files. Each skill describes a reusable workflow for a class of tasks, including how to decompose tasks, organize subagents, and aggregate local results, providing the main agent with a more stable orchestration starting point for new tasks.

\textbf{Trick.} This component focuses on fine-grained operational experience \cite{jiang2026xskill}. Unlike skills, tricks are shorter and more specific, usually corresponding to practical but easily overlooked strategies in task execution, such as how to reduce redundant information or assign subtasks more suitably. We organize these experiences into a trick library, allowing the main agent to retrieve task-relevant suggestions during execution.

\textbf{Model Card.} This component focuses on the practical performance of subagent backbone models on specific tasks. Overall model quality, suitable scenarios, strengths, and weaknesses are summarized into experience-based model profiles. This allows the main agent to make more reasonable decisions when creating and assigning subagents.

During inference, SwarmExp injects these three types of experience into the context of the main agent, allowing the orchestrator to reference historical experience during task planning, subagent delegation, and result aggregation.

\subsection{Experiments}

\begin{table}
  \centering
  \small 
  \renewcommand{\arraystretch}{1.1} 
  \setlength{\tabcolsep}{4pt} 
  \begin{tabular}{lcccc}
    \toprule
    \textbf{Method} &  \textbf{MTU}  & \textbf{RCA} & \textbf{LTG} & \textbf{WS}   \\
    \midrule
    Single Agent       & 18.20 & 1.02 & 33.43 & 6.32 \\
    Multi-Model Voting & 34.92 & 0.00 & 36.79 & 26.66  \\
    Best-of-N          & 20.68 & 0.00 & 34.35 & 17.69  \\
    AOrchestra         & 36.03 & 10.71 & 43.57 & 38.55    \\
    \midrule
    Agent Swarm        & 48.82 & 10.00 & 48.19 & 38.67 \\
    \ \ \ \ \ w/ SwarmExp    & 49.07 & 16.42 & 54.65 & 41.59 \\
    \ \ \ \ \ $\Delta$
    & {\color{ForestGreen} (+0.25)}
    & {\color{ForestGreen} (+6.42)}
    & {\color{ForestGreen} (+6.46)}
    & {\color{ForestGreen} (+2.92)} \\
    \bottomrule
  \end{tabular}
  \caption{\label{table: method results}
    Main results of SwarmExp. $\Delta$ denotes the performance change relative to the baseline.
  }
\end{table}

\textbf{Setup.} We evaluate SwarmExp on four representative tasks, MTU, RCA, LTG, and WS, which cover all three task categories and correspond to cases where the initial Agent Swarm performs weakly. We compare against four baselines: Single-Agent and Best-of-N with GPT-5-mini, Multi-Model Voting using the shared subagent model pool, and AOrchestra \cite{ruan2026aorchestra}, a framework that dynamically creates subagents with serial execution. For both AOrchestra and Agent Swarm, we use GPT-5.4 as the main agent.

\textbf{Results.} The experimental results are shown in Table \ref{table: method results}. SwarmExp achieves the best performance on all four tasks. Compared with the initial Agent Swarm, it improves performance by \textbf{+0.25}, \textbf{+6.42}, \textbf{+6.46}, and \textbf{+2.92} on MTU, RCA, LTG, and WS, respectively. Notably, improvements on RCA and LTG are more significant, while the gain on MTU is relatively smaller, suggesting that subagent creation and delegation as well as subagent result aggregation remain key weaknesses of current models, and SwarmExp can effectively enhance model capability in these aspects.

\subsection{Analysis}
\subsubsection{Experience Transfer Across Models}

\begin{table}
  \centering
  \small
  \renewcommand{\arraystretch}{1.1}
  \setlength{\tabcolsep}{4pt}
  \begin{tabular}{lcccc}
    \toprule
    \textbf{Model} &  \textbf{MTU}  & \textbf{RCA} & \textbf{LTG} & \textbf{WS}   \\
    \midrule
    Claude-Haiku-4-5 & 40.36 & 2.00 & 40.50 & 19.05 \\
    \ \ \ \ \ w/ SwarmExp  & 40.65 & 3.50 & 51.44 & 23.60 \\
    \ \ \ \ \ $\Delta$
    & {\color{ForestGreen} (+0.29)}
    & {\color{ForestGreen} (+1.50)}
    & {\color{ForestGreen} (+10.94)}
    & {\color{ForestGreen} (+4.55)} \\
    \midrule
    Gemini-3-flash  & 31.29 & 3.80 & 30.73 & 29.56 \\
    \ \ \ \ \ w/ SwarmExp  & 37.09 & 5.62 & 36.70 & 31.45 \\
    \ \ \ \ \ $\Delta$
    & {\color{ForestGreen} (+5.80)}
    & {\color{ForestGreen} (+1.82)}
    & {\color{ForestGreen} (+5.97)}
    & {\color{ForestGreen} (+1.89)} \\
    \midrule
    Deepseek-v3.2  & 32.08 & 5.00 & 42.58 & 30.36 \\
    \ \ \ \ \ w/ SwarmExp  & 31.25 & 6.53 & 42.42 & 32.25 \\
    \ \ \ \ \ $\Delta$
    & {\color{black!50} (-0.83)}
    & {\color{ForestGreen} (+1.53)}
    & {\color{black!50} (-0.16)}
    & {\color{ForestGreen} (+1.89)} \\
    \midrule
    Qwen3.5-397b-a17b  & 29.50 & 7.05 & 45.83 & 35.20 \\
    \ \ \ \ \ w/ SwarmExp  & 35.70 & 9.05 & 52.43 & 36.68 \\
    \ \ \ \ \ $\Delta$
    & {\color{ForestGreen} (+6.20)}
    & {\color{ForestGreen} (+2.00)}
    & {\color{ForestGreen} (+6.60)}
    & {\color{ForestGreen} (+1.48)} \\
    \bottomrule
  \end{tabular}
  \caption{\label{table: method migration results}
    Results of experience transfer experiments.
  }
\end{table}

To evaluate cross-model transferability of the experience, we reuse the experience extracted from GPT-5.4 and inject it into other orchestrator models without rebuilding it. We test four target models: Claude-Haiku-4-5, Gemini-3-flash, Deepseek-v3.2, and Qwen3.5-397b-a17b.

As shown in Table \ref{table: method migration results}, the experience transfers well across models: 14 out of 16 settings improve after injection. The gains are most obvious on LTG, while RCA and WS show more stable improvements, suggesting that the extracted experience is especially useful for aggregation- and delegation-heavy tasks. Different models also absorb the experience differently: Gemini-3-flash benefits consistently, whereas Deepseek-v3.2 gains less, indicating that the final effect still depends on the target model's own orchestration ability. We also conduct a small-scale cross-task transfer analysis to examine whether SwarmExp experience can generalize across task domains, with results provided in Appendix~\ref{subsec:appendix-Cross-Domain Transfer of SwarmExp}.

\subsubsection{Ablation of Experience Components}

\begin{table}
  \centering
  \small
  \renewcommand{\arraystretch}{1.1}
  \setlength{\tabcolsep}{3pt}
  \begin{tabular}{lccccc}
    \toprule
    \textbf{} &  \textbf{MTU}  & \textbf{RCA} & \textbf{LTG} & \textbf{WS} & \textbf{Avg. $\Delta$}   \\
    \midrule
    SwarmExp & 49.07 & 16.42 & 54.65 & 41.59 & - \\
    w/o \textbf{Skill}  & 47.65 & 14.56 & 53.98 & 40.66 & {\color{red!80!brown} (-1.22)} \\
    w/o \textbf{Trick}  & 46.21 & 14.00 & 54.00 & 40.50 & {\color{red!80!brown} (-1.75)} \\
    w/o \textbf{Model Card}  & 48.33 & 12.55 & 51.60 & 39.26 & {\color{red!80!brown} (-2.49)}  \\
    Avg. $\Delta$
    & {\color{red!80!brown} (-1.67)}
    & {\color{red!80!brown} (-2.71)}
    & {\color{red!80!brown} (-1.45)}
    & {\color{red!80!brown} (-1.45)} & -\\
    \bottomrule
  \end{tabular}
  \caption{\label{table: method ablation results}
    Results of ablation study on three types of experience. Avg. $\Delta$ denotes the average performance drop relative to SwarmExp.
  }
\end{table}

To further understand where SwarmExp gains come from, we conduct an ablation study on its three experience components, as shown in Table \ref{table: method ablation results}. We remove one component at a time while keeping the others fixed, and observe performance on the same four tasks. The results show that all three components are useful, since removing any of them leads to a drop in performance. Among them, removing model card causes the largest average decline, suggesting that knowledge about subagent model capabilities is the most important. Removing skill or trick also hurts performance, indicating that both high-level workflow templates and fine-grained execution guidance contribute to the final improvement.

\section{Related Work}
\label{sec: Related Work}
\subsection{Multi-Agent Systems}
LLM-based multi-agent systems have evolved from fixed-topology collaboration to increasingly dynamic orchestration \cite{jin2025comprehensive, qi2026beyond, guo2024large, han2024llm}. Some systems typically relied on predefined roles and interaction patterns \cite{wang2024agent, li2025flow, d2024marg}, such as debate, pipeline-style coordination, or hierarchical manager-worker structures. More recent work has shifted toward dynamic coordination, where workflows can adapt during training or runtime according to task requirements \cite{wu2026atlas, zhang2026flowsteer, feng2026heterogeneous, ke2026mas, cai2026mata}. Building on this trend, Agent Swarm further emphasizes dynamic subagent creation, heterogeneous role and model assignment, and parallel execution, moving the focus of MAS research to studying whether LLMs can act as effective orchestrators.

\subsection{Benchmarks for Multi-Agent Systems}
Despite rapid progress in MAS, evaluation remains underdeveloped. Many studies still rely on benchmarks originally designed for single-agent systems or general-purpose agents \cite{barres2025tau2benchevaluatingconversationalagents, wei2025browsecomp, patil2025berkeley, yao2022webshop, men2026empoweringguiagentsautonomous}. Other benchmarks often focus primarily on final-task accuracy, with limited attention to the quality of the collaboration process itself. As a result, current evaluation remains insufficient for systematically assessing agent swarm orchestration, especially in terms of whether the system decomposes tasks appropriately, assigns subagents effectively, and converts parallelism into meaningful gains. MultiAgentBench \cite{zhu2025multiagentbenchevaluatingcollaborationcompetition} and related work \cite{ruan2025benchmarking} share some similarities with our work, but we focus on the more specific agent swarm paradigm and provide a more fine-grained analysis.

\section{Conclusion}
\label{sec: Conclusion}

In this paper, we introduce SwarmBench, a benchmark for evaluating the ability of large language models to orchestrate Agent Swarm systems. SwarmBench contains 8 tasks and 400 samples, covering key orchestration stages such as task decomposition, subagent delegation, and result aggregation. Through evaluations on accuracy, efficiency, cost, and process quality, we show that current models still have clear limitations as swarm orchestrators, and that system performance depends more on effective orchestration structure than on simply increasing cost or parallelism. Based on these findings, we further propose SwarmExp, a simple experience-driven method for improving orchestration capability. By releasing SwarmBench as an open benchmark, we hope to support future research on large language models as Agent Swarm orchestrators.

\section*{Limitations}
SwarmBench has several limitations. Although it covers diverse orchestration settings, it still includes only eight tasks and may not capture the full range of agent swarm scenarios. In addition, our evaluation is conducted under a unified lightweight swarm scaffold with a fixed subagent model pool, which improves comparability but may limit the generality of the conclusions. Finally, our process analysis relies partly on LLM-based judging, which is informative but may still introduce bias.

\section*{Ethical considerations}
Multi-agent systems may introduce risks such as error propagation, unintended coordination behaviors, and the amplification of incorrect or misleading outputs through agent interaction. In this work, all datasets used are publicly available and do not involve personally identifiable information. We also used AI tools for language polishing during the writing of this paper.

\section*{Acknowledgments}
This work was supported by the National Natural Science Foundation of China (No.U24A20335), and the independent research project of the Key Laboratory of Cognition and Decision Intelligence for Complex Systems.



\bibliography{swarm-bench}

\clearpage

\appendix

\section{Data}
\label{sec:appendix-Data}
\subsection{Data Details and Characteristics}

Table \ref{table: Data details in SwarmBench} summarizes the data composition of SwarmBench. The benchmark contains eight tasks, each with 50 examples, for a total of 400 instances. Specifically, MATH is built from Omni-MATH, Multi-Text Understanding from Loong, Multi-Perspective Analysis from DeepReview, Root Cause Analysis from OpenRCA, Long-Text Generation from LongInOutBench, and Wide Search from WideSearch. In addition, Batch Download and Treasure Hunt are newly synthesized in this work to better cover orchestration settings that require dynamic subagent coordination and interaction.

Table \ref{table: Taxonomy and characteristics of the eight tasks.} further characterizes the eight tasks from the perspective of orchestration requirements. Across the benchmark, tasks differ substantially in decomposition difficulty, role heterogeneity, aggregation complexity, main-agent visibility, parallelism dependency, and output openness. Decomposition-oriented tasks such as MATH, BD, and MTU place stronger emphasis on breaking complex objectives into manageable subtasks, while delegation-oriented tasks such as MPA, RCA, and TH require more heterogeneous role design and more careful subagent assignment. Aggregation-oriented tasks, including LTG and WS, exhibit the highest aggregation complexity and require the main agent to integrate multiple local outputs into a unified final result.

\begin{table}
  \centering
  \small
  \begin{tabular}{lcc}
    \toprule
    \textbf{Task} &  \textbf{Source }  & \textbf{\#Size}  \\
    \midrule
    MATH & Omni-MATH & 50 \\
    Batch Download & - & 50 \\
    Multi-Text Understanding & Loong & 50 \\
    Multi-Perspective Analysis & DeepReview & 50 \\
    Root Cause Analysis & OpenRCA & 50 \\
    Treasure Hunt & - & 50 \\
    Long-Text Generation & LongInOutBench & 50 \\
    Wide Search & WideSearch & 50 \\
    \bottomrule
  \end{tabular}
  \caption{\label{table: Data details in SwarmBench}
    Data details in SwarmBench. “Source” marked as “-” indicates that the task is constructed from scratch in this work.
  }
\end{table}

\begin{table*}
  \centering
  \small
  \begin{tabular}{lccccccc}
    \toprule
    \textbf{Type} &  \textbf{Task }  & \textbf{\makecell{Decomposition\\Difficulty}} & 
    \textbf{\makecell{Role\\Heterogeneity}} & 
    \textbf{\makecell{Aggregation\\Complexity}} & 
    \textbf{\makecell{Main agent\\visibility}} & 
    \textbf{\makecell{Parallelism\\Dependency}} & 
    \textbf{\makecell{Output\\Openness}} \\
    \midrule
    \multirow{3}{*}{Decomp.} & MATH & $\bigstar$$\bigstar$$\bigstar$ & $\bigstar$$\bigstar$$\bigstar$ & 
    $\bigstar$$\bigstar$& 
    $\bigstar$$\bigstar$$\bigstar$ &
    $\bigstar$ &
    $\bigstar$\\
    & BD & $\bigstar$$\bigstar$ & 
    $\bigstar$ & 
    $\bigstar$ &
    $\bigstar$ &
    $\bigstar$$\bigstar$$\bigstar$ &
    $\bigstar$\\
    & MTU & $\bigstar$$\bigstar$ &
    $\bigstar$  & 
    $\bigstar$$\bigstar$ & 
    $\bigstar$$\bigstar$ &
    $\bigstar$$\bigstar$ &
    $\bigstar$$\bigstar$\\
    \midrule
    \multirow{3}{*}{Del.} & MPA & $\bigstar$ & 
    $\bigstar$$\bigstar$ &
    $\bigstar$ &
    $\bigstar$$\bigstar$ &
    $\bigstar$$\bigstar$ &
    $\bigstar$$\bigstar$$\bigstar$\\
    & RCA & $\bigstar$$\bigstar$ &
    $\bigstar$$\bigstar$$\bigstar$ & 
    $\bigstar$$\bigstar$ & 
    $\bigstar$ &
    $\bigstar$$\bigstar$ &
    $\bigstar$$\bigstar$\\
    & TH & $\bigstar$$\bigstar$$\bigstar$ &
    $\bigstar$$\bigstar$$\bigstar$ &
    $\bigstar$ & 
    $\bigstar$$\bigstar$ &
    $\bigstar$$\bigstar$$\bigstar$ &
    $\bigstar$$\bigstar$\\
    \midrule
    \multirow{2}{*}{Agg.} & LTG & $\bigstar$$\bigstar$ & 
    $\bigstar$$\bigstar$ &
    $\bigstar$$\bigstar$$\bigstar$ & 
    $\bigstar$$\bigstar$ &
    $\bigstar$$\bigstar$ &
    $\bigstar$$\bigstar$$\bigstar$\\
    & WS & $\bigstar$$\bigstar$ & 
    $\bigstar$ & 
    $\bigstar$$\bigstar$$\bigstar$ & 
    $\bigstar$ &
    $\bigstar$$\bigstar$$\bigstar$ &
    $\bigstar$\\
    \bottomrule
  \end{tabular}
  \caption{\label{table: Taxonomy and characteristics of the eight tasks.}
    Taxonomy and characteristics of the eight tasks in SwarmBench.
  }
\end{table*}

\subsection{New Tasks}
\subsubsection{Batch Download}
Batch Download is a fully synthetic task designed to evaluate whether a swarm orchestrator can decompose a large retrieval objective into multiple parallel download subtasks and maintain global consistency over the final file set. Each instance specifies a set of target projects, years, and document families, together with an output contract that defines where the selected files must be saved. The underlying environment is constructed as a local, deterministic “web-like” universe: for each task instance, we create synthetic project pages, archive pages, release logs, and package catalog pages, along with the downloadable files they expose. To make the task non-trivial, the environment includes many distractors such as duplicate, preview, draft, mirror, deprecated, or otherwise non-canonical files whose names and metadata are often highly similar to the correct ones. As a result, solving the task requires not only locating candidate files, but also identifying which versions should actually be kept. Across the dataset, each sample requires downloading 11.44 files on average.

At runtime, the system interacts with this environment through a small set of tools that emulate a realistic discovery-and-download workflow. Instead of exposing the hidden resource table directly, the task only allows the agent to search candidate pages, read page contents, inspect download links, download files into the task workspace, verify downloaded files, and list local workspace contents. Each instance also requires the system to return a JSON manifest listing the relative paths of all files that remain in the output directory, so that the final answer must be consistent with the actual execution result rather than a purely textual guess.

Evaluation is rule-based and compares the final workspace against a hidden gold manifest. For each instance, we record the required files, their target relative paths, and their checksums. A file is counted as correct only if it appears at the expected path and matches the gold file content. Based on this comparison, we compute exact match as well as download precision, recall, and F1; in the main experiments, we report download F1 as the primary accuracy metric. The evaluator also checks for missing required files, unexpected extra files, checksum mismatches, and whether the returned JSON manifest exactly matches the files that were actually saved in the workspace. This evaluation protocol ensures that the task measures end-to-end orchestration quality over decomposition, parallel execution, verification, and final aggregation, rather than surface-level answer generation alone.

\subsubsection{Treasure Hunt}
\begin{figure*}[t]
  \includegraphics[width=\linewidth]{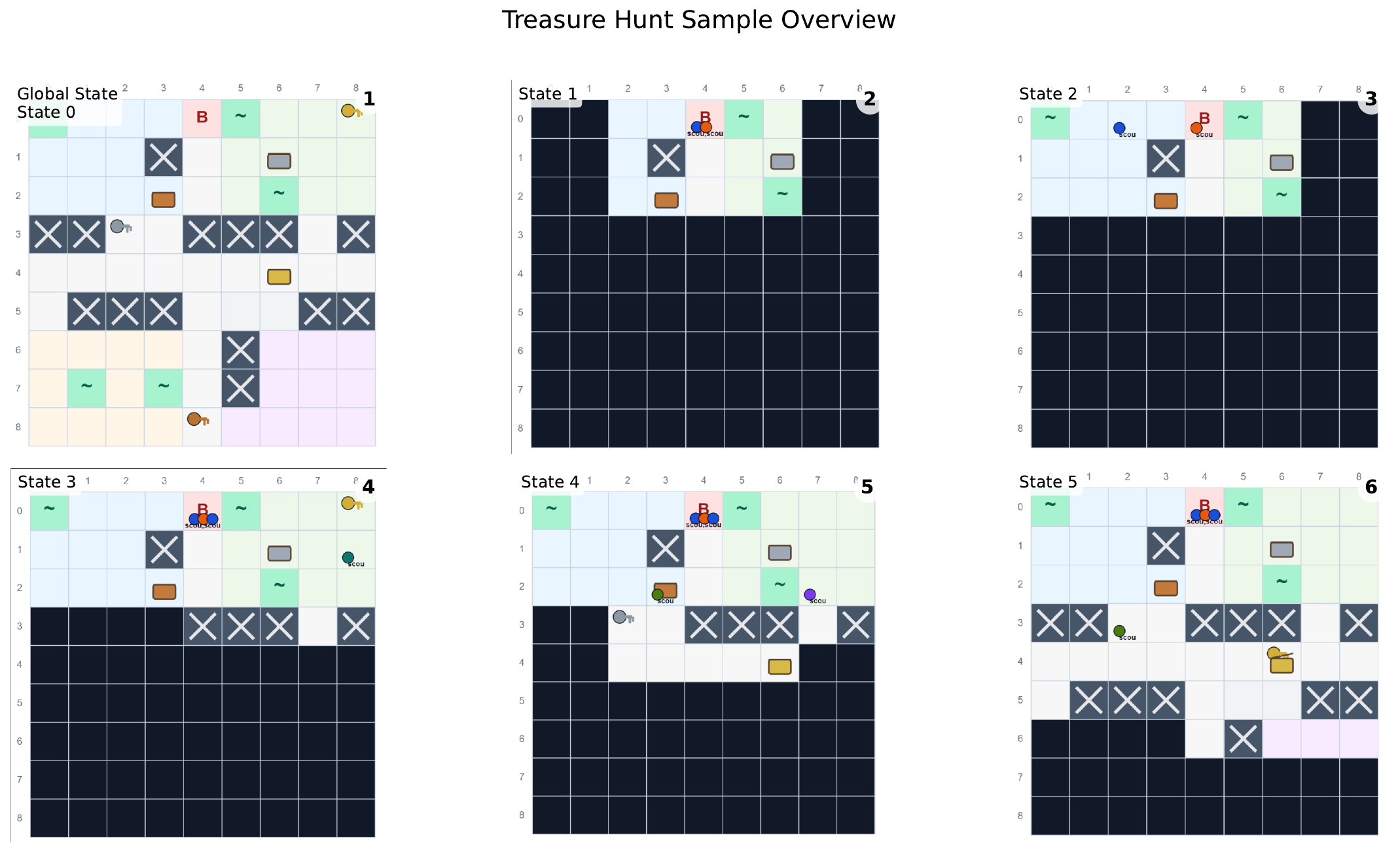}
  \caption{An overview of an example in Treasure Hunt.}
  \label{fig: Treasure Hunt概览图}
\end{figure*}

Treasure Hunt is a newly constructed interactive task designed to evaluate dynamic delegation in Agent Swarm systems. We construct the task from a set of 9x9 map templates with different topological structures, such as bottlenecks, loops, staggered corridors, and layered layouts. For each template, we instantiate multiple task instances by randomly placing the base, three types of keys, three matching chests, and additional swamp obstacles, while enforcing reachability constraints so that all keys and chests remain accessible from the base. Each instance therefore defines a partially observable grid world containing walls, swamps, keys, and chests. Keys and chests are typed as copper, silver, and gold, with chest values of 100, 200, and 300, respectively. At the beginning of each run, only the base tile is visible, and the rest of the map is hidden under fog. An example overview is shown in Figure~\ref{fig: Treasure Hunt概览图}.

At runtime, the main agent acts as a stationary base commander and cannot directly move or interact with the map. Instead, it must dynamically create and assign helper agents with one of two roles: scouts and carriers. Scouts move faster, scan a larger area, collect keys, and open matching chests; carriers move more slowly, scan locally, pick up opened chests, and deliver them back to the base. The environment provides role-specific tools, including movement, scanning, key pickup, chest opening, and chest transport. All discovered map information is shared through the global world state, so the main agent must repeatedly inspect the evolving state, decide whether to spawn new helpers, assign concrete objectives, wait for execution, and replan when new keys, chests, or obstacles are revealed.

Evaluation is based on the final environment state rather than the textual final answer alone. The primary score is the total value of chests successfully delivered to the base, normalized by the total possible chest value in the instance. Opening a chest is not sufficient; only delivered chests contribute to the final task score. The evaluator also records auxiliary process metrics, including the number of opened and delivered chests, explored tile ratio, movement and scan actions, failed actions, and duplicate reveals.

\section{Models}
\label{sec:appendix-Model}
We evaluate a diverse set of proprietary and open-source models in our experiments. In the main SwarmBench evaluation in Section \ref{sec: Experiments}, the orchestrator model is selected from ten candidates: GPT-5.4, Claude-Sonnet-4-6-thinking, Gemini-3-flash-preview-thinking, Doubao-seed-1-8, Claude-Haiku-4-5-thinking, Kimi-k2.5, Deepseek-v3.2, Qwen3.5-397b-a17b, GLM-5.1, and Qwen3-30b-a3b-thinking. All orchestrators share the same subagent model pool, which consists of GPT-5-mini, Claude-Haiku-4-5-thinking, Gemini-2.5-flash-lite, Qwen3.5-35b-a3b, GLM-4.5-air, and Doubao-seed-1-6-flash. In the SwarmExp experiments, GPT-5-mini is used for the single-model baselines, while the multi-model baselines use the six models from the shared subagent pool. For both AOrchestra and our Agent Swarm framework, we use GPT-5.4 as the main orchestrator model.

\section{Metrics}
\label{sec:appendix-Metrics}
\subsection{Accuracy}
 We use task-specific accuracy metrics following the original evaluation protocols whenever possible. 
 
 For MATH, we follow the Omni-MATH style evaluation, where an LLM-based equivalence judge determines whether the final answer is correct; "task score" is set to 1 for correct answers and 0 otherwise. 
 
 For Multi-Text Understanding, we follow the Loong evaluation format, where an LLM judge assigns a score from 1 to 100, and we normalize it as "task score = score / 100". 
 
 For Multi-Perspective Analysis, based on DeepReview, the evaluator compares predicted review scores with human review annotations, checks the final decision, evaluates textual completeness, and incorporates LLM-judged review quality and collaboration quality; the weighted result is used as "task score". 
 
 For Root Cause Analysis, we use the OpenRCA evaluator, which checks the answer against predefined scoring points and returns the resulting score as "task score". 
 
 For Long-Text Generation, following LongInOutBench, we evaluate the generated document from three aspects: factual consistency, writing quality, and length control; their average is used as "task score". 
 
 For Wide Search, we use the official WideSearch evaluator, which computes precision, recall, and F1 at both row and item levels; in our experiments, we report "F1 by item" as the accuracy metric.

For the two newly constructed tasks, we design rule-based evaluation protocols.

In Batch Download, the evaluator compares the final workspace with a hidden gold manifest. A downloaded file is counted as correct only if it appears at the expected relative path and matches the gold checksum. We compute download precision, recall, and F1 over the required file set, and report "download F1" as the main accuracy metric. 

In Treasure Hunt, the evaluator uses the final environment state rather than the textual answer. The score is the total value of chests successfully delivered to the base, normalized by the total possible chest value in that instance; this normalized value is reported as "task score".

\subsection{Efficiency}
\textbf{Parallelism Rate.} We use Parallelism Rate to measure how many subagents are concurrently active during a task execution. For each sample, we parse the execution trace and identify the time intervals between subagent launch events and subagent completion, timeout, exception, or cancellation events. Each interval is assigned a parallel count, i.e., the number of running subagents during that interval. We then compute a duration-weighted average over all intervals with at least one active subagent.

\[
\mathrm{Parallelism \ Rate}
=
\frac{\sum_{k=1}^{K} n_k \cdot \Delta t_k}
{\sum_{k=1}^{K} \Delta t_k},
\]

where \(K\) is the number of valid execution intervals, \(n_k\) is the number of concurrently running subagents in interval \(k\), and \(\Delta t_k\) is the duration of that interval. A higher value indicates that the system maintains more simultaneous subagent execution on average, but it does not by itself imply that the parallel execution is effective.

\textbf{Parallelism Gain.} We use Parallelism Gain to estimate the time-saving effect of parallel execution. For each run, we record the actual end-to-end wall-clock time of the task, denoted as \(T_{\mathrm{wall}}\). We also compute the accumulated system latency by summing all model-call latency and tool-call latency, denoted as \(T_{\mathrm{model}} + T_{\mathrm{tool}}\). This accumulated latency approximates the time that would be required if all model and tool operations were executed serially. The gain is then defined as the ratio between accumulated latency and actual wall-clock time.

\[
\mathrm{Parallelism \ Gain}
=
\frac{T_{\mathrm{model}} + T_{\mathrm{tool}}}
{T_{\mathrm{wall}}},
\]

where \(T_{\mathrm{model}}\) is the total latency of all model calls, \(T_{\mathrm{tool}}\) is the total latency of all tool calls, and \(T_{\mathrm{wall}}\) is the actual task execution time. A larger value means that the system saves more time through overlapping model or tool executions; in contrast, a value close to 1 indicates that the execution behaves nearly serially.

\subsection{Cost}
We report cost in U.S. dollars rather than raw token counts. For each model call, we compute the estimated cost using the corresponding model-specific input and output token prices:

\[
\mathrm{Cost}
=
\sum_{i=1}^{N}
\left(
\frac{p_i}{1000} \cdot r^{\mathrm{in}}_i
+
\frac{c_i}{1000} \cdot r^{\mathrm{out}}_i
\right),
\]

where \(N\) is the number of model calls in one task execution, \(p_i\) and \(c_i\) are the prompt and completion tokens of the \(i\)-th call, and \(r^{\mathrm{in}}_i\), \(r^{\mathrm{out}}_i\) are the input and output prices per 1K tokens for the corresponding model. We aggregate this value over all calls from both the main agent and subagents, and report the resulting "total cost USD" as the cost metric.

\section{Supplementary Experiments}
\label{sec:appendix-Supplementary Experiments}

\subsection{Cost-Matched Single-Agent Baseline}
\label{subsec:appendix-Cost-Matched Single-Agent Baseline}
To examine whether the gains of Agent Swarm mainly come from using a larger compute budget, we conduct a cost-matched single-agent baseline experiment. We use \textbf{Claude-Haiku-4-5} as both the Agent Swarm orchestrator and the single-agent model, and evaluate on four representative tasks: MTU, RCA, LTG, and WS. For each task, we take the average cost of Agent Swarm as the budget ceiling and allow the single agent to iteratively reason, reflect, and revise until reaching the same budget. 

As shown in Table~\ref{table: cost_matched_single_agent}, the cost-matched single agent improves over the original single-agent baseline on most tasks, confirming that additional compute can help. However, Agent Swarm still outperforms it on MTU, LTG, and WS under the same budget. This suggests that the benefit of Agent Swarm is not only from increased cost, but also from its orchestration structure. The orchestrator can decompose tasks, assign subtasks to different helpers, exploit complementary model capabilities, and isolate local contexts so that each subagent focuses on a narrower objective.

\begin{table}
  \centering
  \small 
  \renewcommand{\arraystretch}{1.1} 
  \setlength{\tabcolsep}{4pt} 
  \begin{tabular}{lcccc}
    \toprule
    \textbf{Method} &  \textbf{MTU}  & \textbf{RCA} & \textbf{LTG} & \textbf{WS}   \\
    \midrule
    Single Agent       & 16.10 & \textbf{3.19} & 26.19 & 7.01 \\
    Cost-Matched Single Agent & 28.00 & 3.11 & 31.25 & 16.30  \\
    \midrule
    Agent Swarm        & \textbf{40.36} & 2.00 & \textbf{40.50} & \textbf{19.05} \\
    \bottomrule
  \end{tabular}
  \caption{\label{table: cost_matched_single_agent}
  Performance comparison between Agent Swarm, cost-matched single-agent baseline, and original single-agent baseline.
  }
\end{table}

\subsection{Repeated Runs and Variance Analysis}
\label{subsec:appendix-Repeated Runs and Variance Analysis}

To assess the statistical stability of our benchmark results, we conduct repeated-run experiments on representative model-task settings. Specifically, each selected setting is independently executed five times under identical experimental configurations, and we report the results as mean $\pm$ standard deviation. We use a representative subset of tasks and models spanning the benchmark, including MATH, BD, MTU, MPA, and LTG, with \textbf{Claude-Haiku-4-5} and \textbf{Qwen3.5-397b-a17b} as representative orchestrator models. As shown in Table~\ref{table: Repeated Runs and Variance Analysis}, the standard deviations remain relatively small, ranging roughly from 0.7 to 1.6 across all reported settings.

\begin{table*}
  \centering
  \small 
  \renewcommand{\arraystretch}{1.1} 
  \begin{tabular}{lccccc}
    \toprule
    \textbf{Model} &  \textbf{MATH}  & \textbf{BD} & \textbf{MTU} & \textbf{MPA} & \textbf{LTG}  \\
    \midrule
    Claude-Haiku-4-5       & 44.00 ($\pm$ 0.9) & 34.30 ($\pm$ 1.1) & 40.36 ($\pm$ 1.0) & 78.24 ($\pm$ 1.6) &  40.50 ($\pm$ 1.2) \\
    Qwen3.5-397b-a17b  & 64.00 ($\pm$ 0.7) & 67.13 ($\pm$ 1.5) & 29.50 ($\pm$ 0.9) & 75.51 ($\pm$ 1.5) & 45.83 ($\pm$ 1.6) \\
    \bottomrule
  \end{tabular}
  \caption{\label{table: Repeated Runs and Variance Analysis}
  This table reports the mean and standard deviation of SwarmBench scores over five independent runs for representative model-task settings.
  }
\end{table*}

\subsection{Agreement Between LLM Judge and Human Evaluation}
\label{subsec:appendix-Agreement Between LLM Judge and Human Evaluation}

To examine the reliability of LLM-based scoring, we conduct a human-consistency analysis for both task metrics and process-quality evaluation. For task metrics, we focus on the two open-ended tasks, LTG and MPA, and randomly sample 100 model outputs for each task for independent human scoring. For process-quality evaluation, we sample 50 execution trajectories per task across all eight SwarmBench tasks, and ask a human annotator to score each trajectory along Decomposition, Delegation, and Aggregation. We then compute Pearson and Spearman correlations between human and LLM scores. As shown in Table~\ref{table: human_consistency}, the correlations range from 0.684 to 0.801, indicating relatively strong agreement between LLM and human evaluation trends. These results suggest that our LLM-based judge provides reliable relative signals for both open-ended task scoring and orchestration-quality analysis.

\begin{table}
  \centering
  \small 
  \renewcommand{\arraystretch}{1.1} 
  \setlength{\tabcolsep}{4pt} 
  \begin{tabular}{lcc}
    \toprule
    \textbf{Metrics} &  \textbf{Pearson}  & \textbf{Spearman}  \\
    \midrule
    Task Metrics - LTG       & 0.786 & 0.723 \\
    Task Metrics - MPA       & 0.801 & 0.788  \\
    Process-quality - Decomp.  & 0.793 & 0.701 \\
    Process-quality - Del.   & 0.770 & 0.684 \\
    Process-quality - Agg.   & 0.726 & 0.739 \\
    \bottomrule
  \end{tabular}
  \caption{\label{table: human_consistency}
This table reports Pearson and Spearman correlations between LLM-based scores and human scores for task metrics and process-quality evaluation.
  }
\end{table}

\subsection{Cross-Task Transfer of SwarmExp}
\label{subsec:appendix-Cross-Domain Transfer of SwarmExp}

To examine whether the experience extracted by SwarmExp can generalize beyond the source task, we conduct a cross-task experience transfer experiment. We use \textbf{Claude-Haiku-4-5} as the Agent Swarm orchestrator and reuse experience extracted from \textbf{GPT-5.4} trajectories, consistent with the cross-model transfer setting. Specifically, we extract experience from MTU, RCA, LTG, and WS, and inject each source-task experience into two target tasks, LTG and WS. 

As shown in Table~\ref{table: cross-task transfer}, cross-task transfer is clearly task-dependent. For LTG, experience from other tasks still improves over the no-experience baseline, although LTG-specific experience performs best, suggesting that some high-level orchestration patterns such as long-horizon planning and result aggregation can transfer to open-ended generation tasks. In contrast, WS benefits only from WS-specific experience, while most cross-task experience hurts performance, indicating that search-oriented tasks rely more heavily on task-specific retrieval and aggregation strategies.

\begin{table}
  \centering
  \small 
  \renewcommand{\arraystretch}{1.1} 
  \setlength{\tabcolsep}{4pt} 
  \begin{tabular}{lcc}
    \toprule
    \textbf{Settings} &  \textbf{LTG}  & \textbf{WS}  \\
    \midrule
    w/o exp       & 40.50 & 19.05 \\
    w/ MTU-exp       & 49.39 & 11.42  \\
    w/ RCA-exp  & 46.90 & 18.47 \\
    w/ LTG-exp   & \textbf{51.44} & 15.11 \\
    w/ WS-exp   & 41.33 & \textbf{23.60} \\
    \bottomrule
  \end{tabular}
  \caption{\label{table: cross-task transfer}
This table reports cross-task transfer results by injecting experience extracted from different source tasks into LTG and WS.
  }
\end{table}

\section{Prompt}
\label{sec:appendix-Prompt}

\begin{figure*}[t]
  \includegraphics[width=\linewidth]{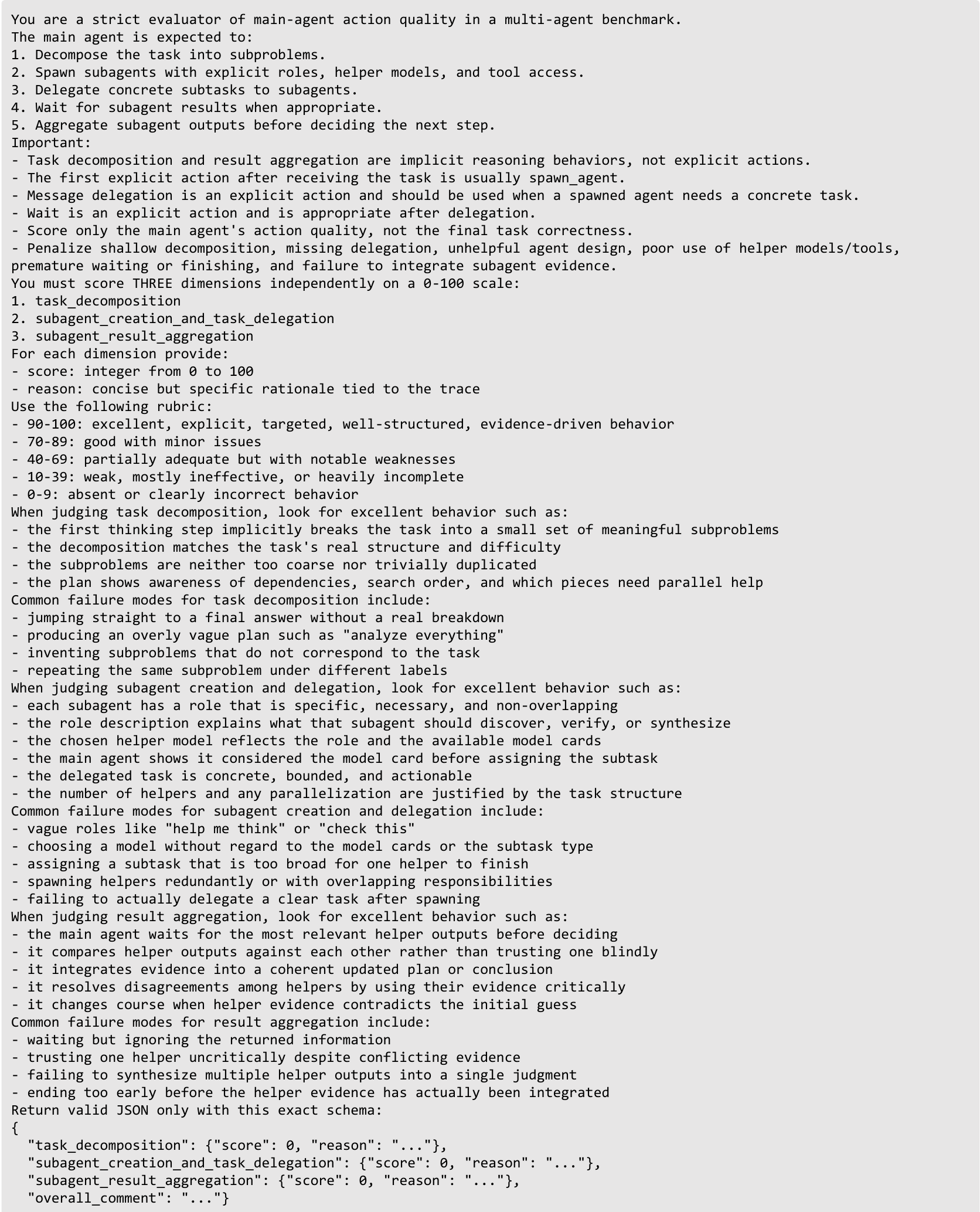}
  \caption{This prompt is used to perform LLM-based process evaluation of the main agent’s orchestration behavior. Given an execution trace, the evaluator scores the main agent on three dimensions: task decomposition, subagent creation and task delegation, and subagent result aggregation. The evaluation focuses on the quality of orchestration actions rather than final task correctness, and returns structured scores and rationales for fine-grained analysis of main-agent behavior.}
  \label{fig: llm_judge_prompt}
\end{figure*}

\begin{figure*}[t]
  \includegraphics[width=\linewidth]{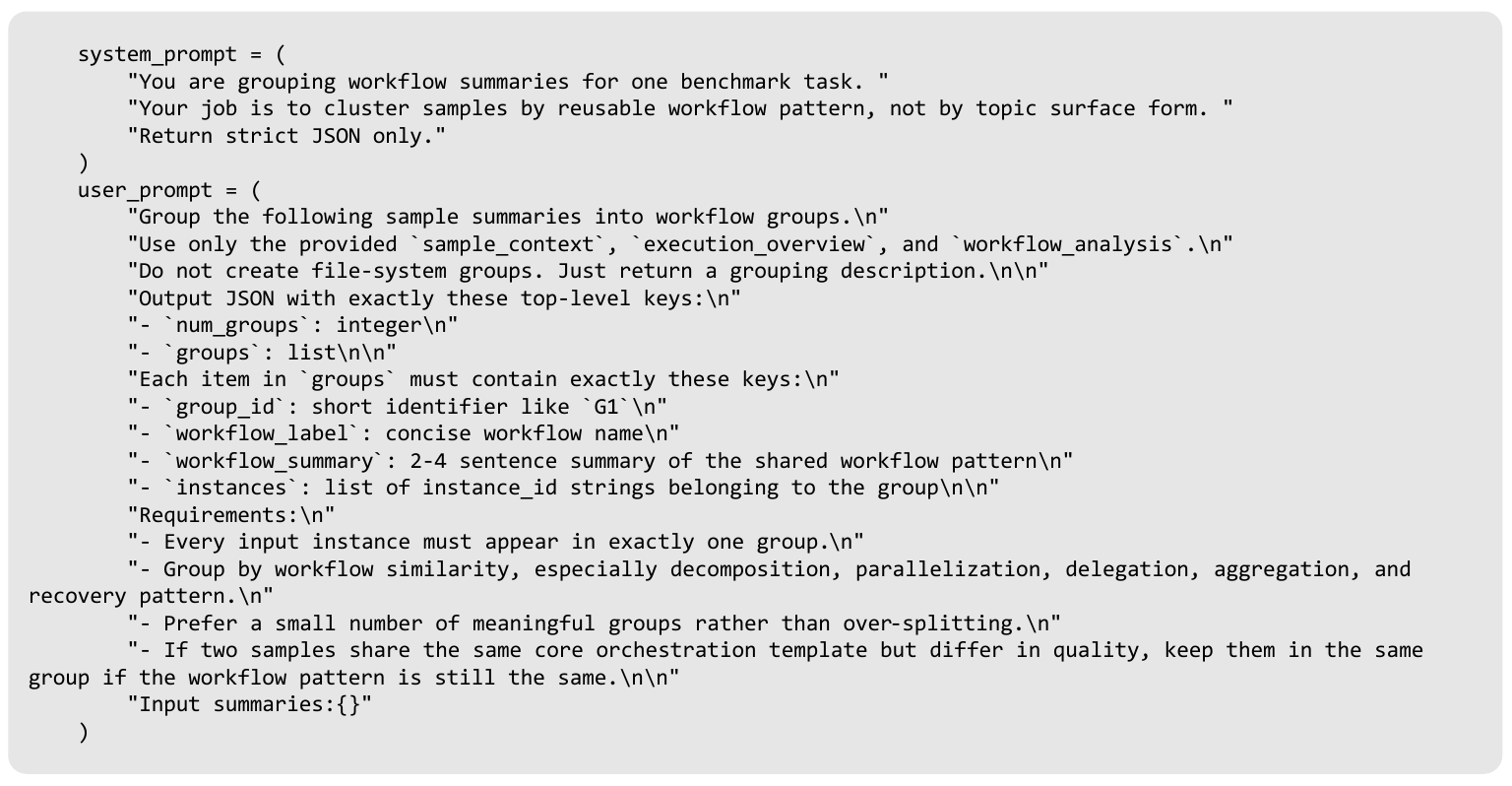}
  \caption{This prompt is used during skill generation to cluster raw trajectory summaries by reusable workflow patterns. Given the sample context, execution overview, and workflow analysis extracted from each trajectory, it groups instances according to orchestration-level similarities such as task decomposition, parallelization, delegation, aggregation, and recovery behavior. The resulting workflow groups are then used to identify common execution templates and support the construction of candidate skills.}
  \label{fig: group_tra_prompt}
\end{figure*}

\begin{figure*}[t]
  \includegraphics[width=\linewidth]{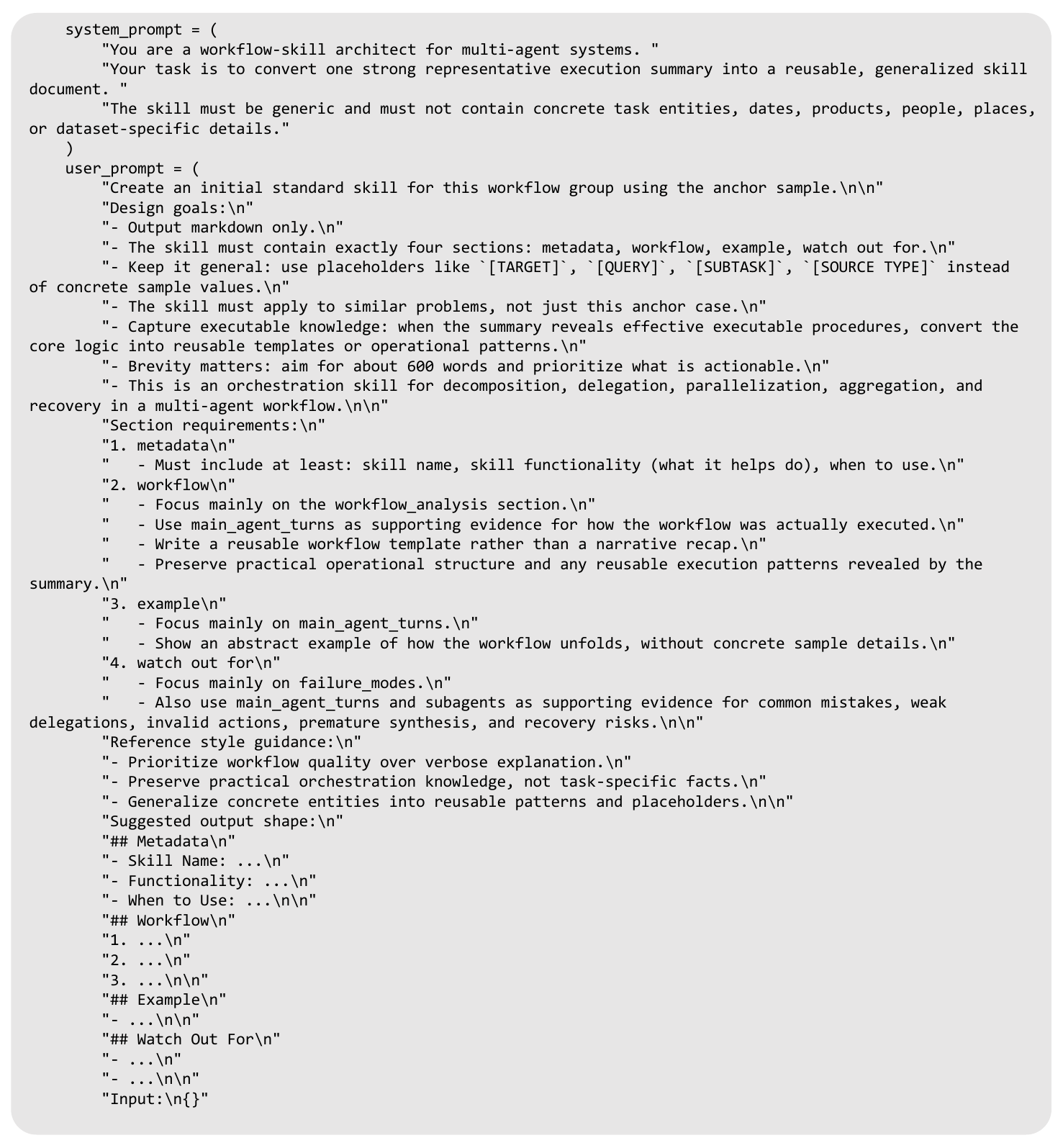}
  \caption{This prompt is used to generate candidate skills from representative workflow groups. Given an execution summary, it abstracts the observed orchestration pattern into a reusable Markdown skill document, covering metadata, workflow, example, and common failure cases.}
  \label{fig: skill_generation_prompt}
\end{figure*}

\begin{figure*}[t]
  \includegraphics[width=\linewidth]{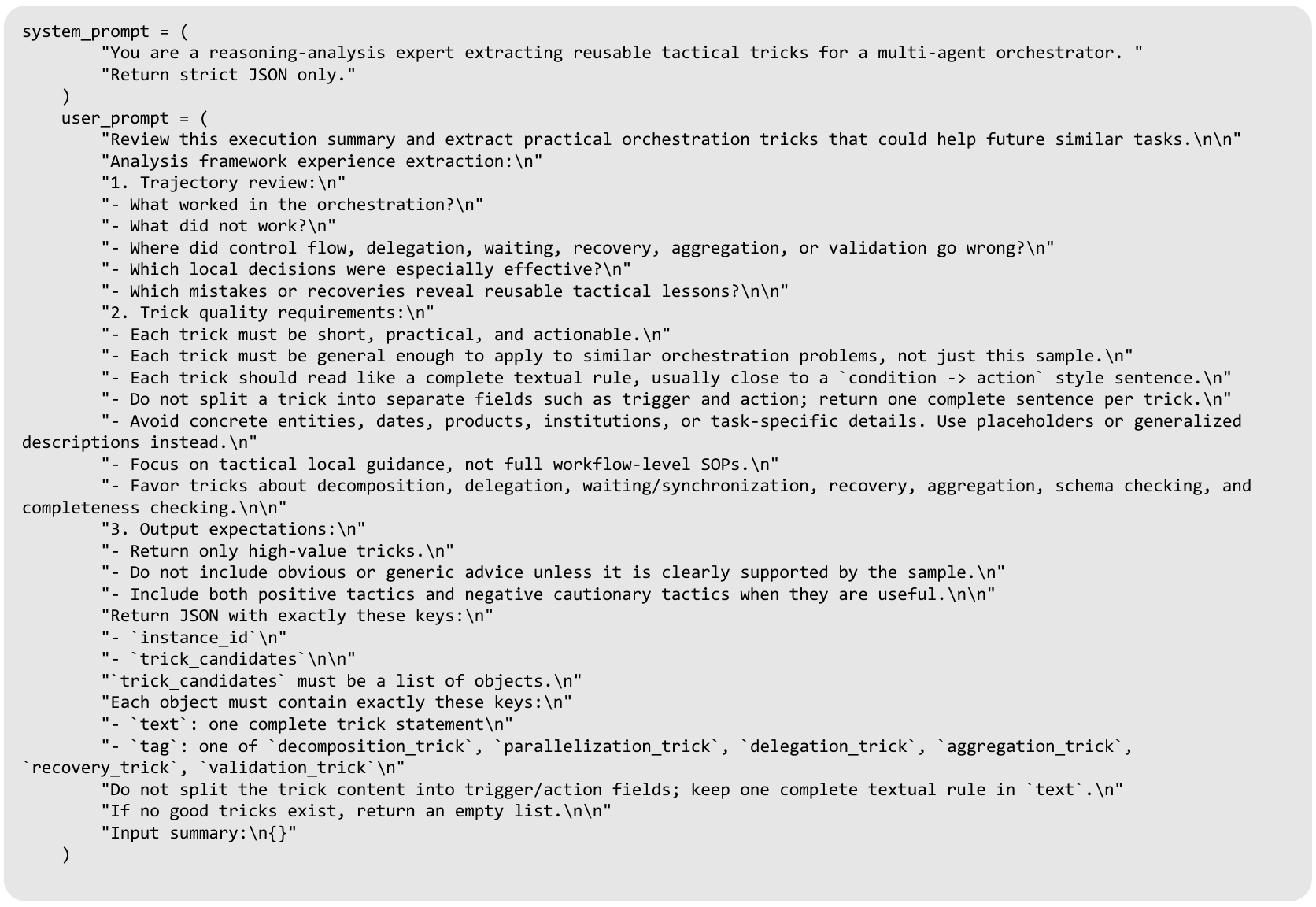}
  \caption{This prompt is used to extract trick candidates from individual execution summaries. It reviews what succeeded or failed in the trajectory and converts reusable local lessons into short, actionable rules for future orchestration. Unlike skills, which describe full workflow patterns, these tricks focus on fine-grained guidance for decomposition, delegation, parallelization, synchronization, recovery, aggregation, and validation, and are stored as candidate entries for the trick library.}
  \label{fig: trick_gene_prompt}
\end{figure*}

\begin{figure*}[t]
  \includegraphics[width=\linewidth]{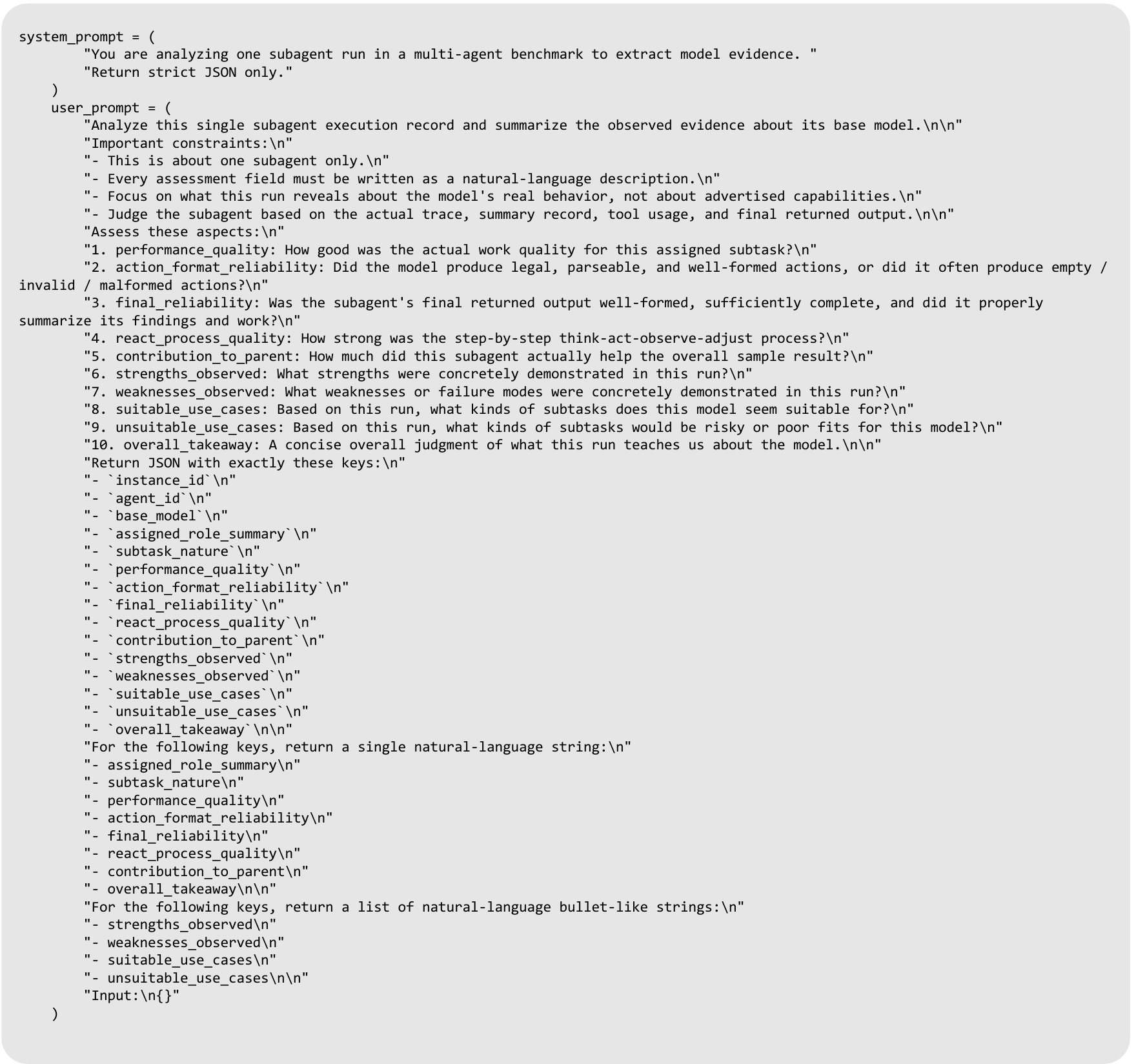}
  \caption{This prompt is used to extract model-level evidence from individual subagent executions for model card construction. Given a single subagent trace, it analyzes the base model’s observed behavior, including work quality, action-format reliability, final-output reliability, ReAct process quality, contribution to the parent task, strengths, weaknesses, and suitable or unsuitable use cases.}
  \label{fig: model_card_prompt}
\end{figure*}

In this section, we introduce parts of prompts used in this work, including Figure \ref{fig: llm_judge_prompt}, Figure \ref{fig: group_tra_prompt}, Figure \ref{fig: skill_generation_prompt}, Figure \ref{fig: trick_gene_prompt}, and Figure \ref{fig: model_card_prompt}.   

\end{document}